\documentclass[11pt,a4paper]{article}
\usepackage{times,latexsym}
\usepackage{url}
\usepackage[T1]{fontenc}

\usepackage{microtype}
\usepackage{graphicx}
\usepackage{subfigure}
\usepackage{booktabs} 

\usepackage{amsmath}
\usepackage{amssymb}
\usepackage{mathtools}
\usepackage{amsthm}

\usepackage{url}
\usepackage{times}
\usepackage{latexsym}
\usepackage{subcaption}
\usepackage{bm}
\usepackage{multirow}
\usepackage{dsfont}
\usepackage{xcolor}
\usepackage{colortbl}
\usepackage{microtype}
\usepackage{adjustbox}
\usepackage{enumitem}
\usepackage{makecell}

\usepackage{verbatim} 

\usepackage[acceptedWithA]{tacl2021v1}

\usepackage{xspace,mfirstuc,tabulary}

\newif\iftaclinstructions
\taclinstructionsfalse 
\iftaclinstructions
\renewcommand{\confidential}{}
\renewcommand{\anonsubtext}{(No author info supplied here, for consistency with
TACL-submission anonymization requirements)}
\newcommand{\instr}
\fi

\iftaclpubformat 

\else

\fi

\definecolor{red}{rgb}{0.635, 0.0, 0.145}
\definecolor{blue}{rgb}{0.0, 0.314, 0.937}
\definecolor{gray}{rgb}{0.96,0.96,0.98}
\newcolumntype{g}{>{\columncolor{gray}}c}

\title{Post-hoc Disentanglement of Textual and Acoustic Features \\in Pre-trained Speech Encoders}

\author{
  Hosein Mohebbi$^{1,2}$ ~ Grzegorz Chrupa\l{}a$^1$ ~ Willem Zuidema$^2$ ~ Afra Alishahi$^1$ ~ Ivan Titov$^{2,3}$\\
  $^1$Tilburg University ~
  $^2$University of Amsterdam ~
  $^3$University of Edinburgh \\
  \hspace{5pt}\texttt{\{h.mohebbi, w.h.zuidema\}@uva.nl}, \hspace{5pt}\texttt{grzegorz@chrupala.me},\\ \texttt{a.alishahi@tilburguniversity.edu, ititov@inf.ed.ac.uk}
}

\date{}

\begin{document}

\maketitle

\begin{abstract}
Pre-trained speech models build entangled internal representations, which capture a variety of features (e.g., pitch, loudness, syntax, or semantics of an utterance) in a distributed encoding. This entanglement makes it difficult to track how such representations rely on textual versus acoustic information when used in downstream applications, limiting their interpretability and transparency.
In this paper, we explore an approach to post-hoc disentanglement that 
builds on the Information Bottleneck principle. We take representations from pre-trained speech encoders as input, and separate them into two distinct components in a cascaded fashion: first we learn a vector encoding content (i.e., what can be transcribed as text), and nothing else; second, we learn a vector encoding all complementary acoustic features relevant to a target task. 
We apply and evaluate our proposed setup to emotion recognition and speaker identification target tasks, quantifying the relative contribution of textual and acoustic features at each model layer.
Finally, we use the approach for disentangled feature attribution as it allows us to identify the most salient speech frames from both the textual and acoustic perspectives.\footnote{Code is publicly released at \url{https://github.com/hmohebbi/disentangling_representations}.}
\end{abstract}

\section{Introduction}
\label{sec:introduction}

\begin{figure}[t]
    \centering
    \includegraphics[width=\columnwidth]{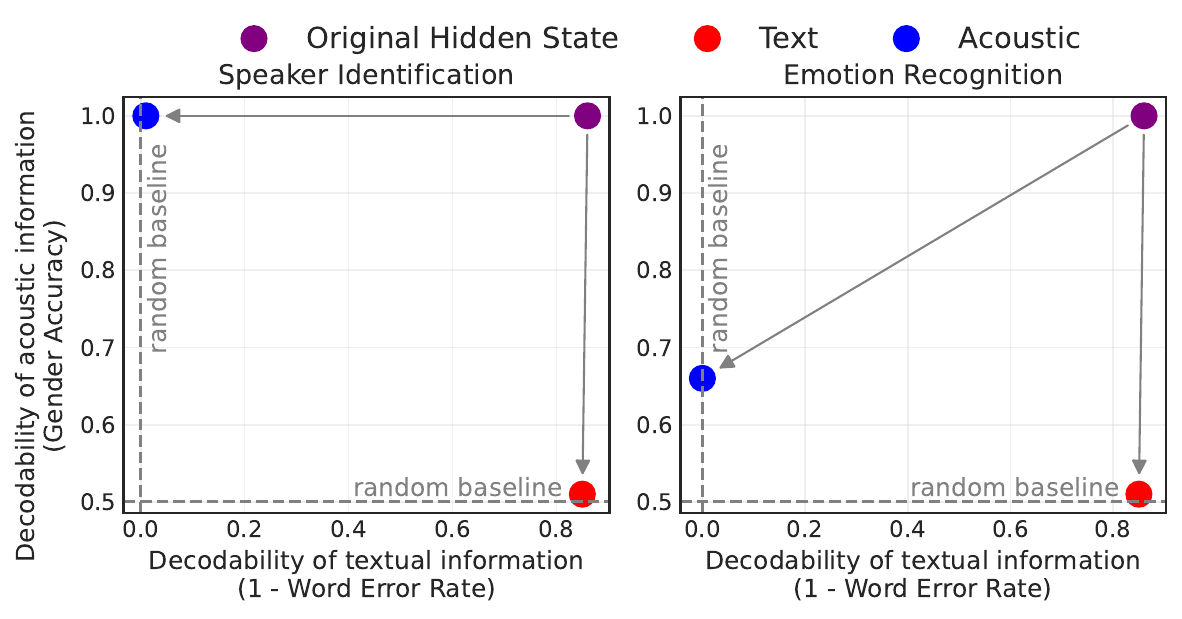}
    \caption{We take the original hidden states from a pre-trained speech model and separate them into \textcolor{red}{text} and \textcolor{blue}{acoustic} components. The textual component achieves ASR performance on par with the original hidden states (measured by WER), yet reduces the decodability of acoustic features (e.g., speaker gender) to chance levels. Conversely, the acoustic component exhibits random WER while effectively predicting gender, with its decodability varying depending on the target task (thus satisfying our third desideratum). Scores are averaged over the test set; see Table \ref{tab:sanity_check} for exhaustive results.}
    \label{fig:eval_intro}
\end{figure}

Speech is a natural and expressive input modality for training and interacting with language models.
Beyond conveying semantic content, it carries paralinguistic cues that enrich context and help the model better infer user intent.
In many downstream speech applications, modern AI systems rely on pre-trained speech encoders such as Wav2Vec2 \citep{baevski2020wav2vec}, HuBERT \citep{Hsu2021HuBERTSS}, and Whisper \citep{Radford2022RobustSR}. Such models produce speech representations that are highly \emph{entangled}. 
This means that distinct characteristics (e.g.\ pitch, loudness, syntax, and semantics) of a spoken utterance are not separated into specialized dimensions within the model's latent space, but are instead mixed together. 
Entanglement is a major obstacle to interpretability and controllability; it makes it difficult to track how models actually use textual versus acoustic information when used in downstream applications.
\emph{Disentanglement}, to the extent that it is possible, is therefore often highly desirable. 

Learning disentangled representations such that distinct factors are separable and controllable has been widely studied \cite{wang2024disentangledrepresentationlearning}. The literature (e.g., \citealt{bengio2013representation}) provides many arguments for its importance in deep learning. 
In the speech domain specifically, disentanglement has been used for controllable style transfer in voice conversion, where the goal is to modify para-linguistic information while preserving linguistic content \citep{NIPS2017_7a98af17, polyak2021speech, Qian2019ZeroShotVS, Qian2020UnsupervisedSD, Choi2021NeuralAA, Choi2022NANSYUV, Qu_2024}.
What is missing is a general-purpose disentanglement approach that satisfies the following desiderata:
\begin{itemize}[itemsep=0pt, topsep=0pt]
    \item The approach must disentangle representations learned by pre-existing, pre-trained models, with minimal data or computation. 
    Since many downstream speech applications rely on pre-trained speech encoders as starting points, the ability to perform post-hoc disentanglement of their representations is essential for both interpretability and controllability.
    Methods such as AutoVC~\citep{Qian2019ZeroShotVS}, SpeechSplit~\citep{Qian2020UnsupervisedSD}, Prosody2Vec~\citep{Qu_2024}, and FACodec~\citep{Ju2024NaturalSpeech3Z} learn disentangled representations from scratch, and even sometimes require large amounts of fine-tuning data. For example, training Prosody2Vec takes around three weeks \citep{Qu_2024}, as parameters are first pre-trained on spontaneous and emotional raw speech and then fine-tuned for emotion recognition as the target task.
    \item The approach must be general and applicable to models with varying architectures, sizes, and learning objectives. Many existing methods are tailored to a specific architecture; for example, Prosody2Vec \citep{Qu_2024} builds on top of HuBERT's \citep{Hsu2021HuBERTSS} quantization module to encode content.
    \item The approach should allow for acoustic features to emerge based on their relevance to a target downstream task, rather than being based on pre-specified components such as pitch and timbre (as in e.g.\ SpeechSplit \citealp{Qian2020UnsupervisedSD}). This is important, as defining acoustic dimensions in advance might result in losing task-relevant aspects of acoustic features.
\end{itemize}

In this paper, we present a post-hoc approach to disentangle pre-trained speech representations into textual and task-relevant acoustic components, satisfying the proposed desiderata.
Our contributions are threefold. 
Firstly, we show that our proposed approach, based on the Variational Information Bottleneck \citep{tishby2000information} and cascaded disentanglement (instead of joint optimization) delivers the effective, and generally-applicable approach that we are after.
In section~\ref{sec:approach} we describe this setup in detail, and in Appendix \ref{app:ablation}, we report two ablation experiments that show both the bottleneck and the cascading are crucial for these results.

Secondly, in Section~\ref{sec:evaluation}, we comprehensively evaluate our method and related work to assess the extent to which the resulting textual and acoustic components are in fact disentangled and contain minimal overlapping linguistic information. 
Figure~\ref{fig:eval_intro} illustrates the core idea.
Our probing experiments show that textual latent representations can predict transcriptions as effectively as the original speech representations, but exhibit random performance for predicting acoustic features such as intensity, pitch, or speaker gender and identity. In contrast, acoustic representations excel at predicting these features (and vary for different target tasks) but show random performance for transcriptions, validating the effectiveness of our disentanglement approach.  

Thirdly, we leverage our approach for improved interpretability.
In section \ref{sec:layerwise_contribution}, we trace the evolution of textual versus acoustic aspects of the speech representations across layers of the original models. 
Further, in Section \ref{sec:feature_attribution}, we demonstrate our approach can also serve as a \emph{disentangled} feature attribution method to quantify the impact of individual frames in the input speech signal to a given target task. 
Prior work in this area has highlighted that a key challenge in feature attribution for the speech domain lies in identifying informative features \citep{fucci-etal-2024-explainability, Shen2025OnTR}.
Unlike existing gradient-based methods \citep{10.5555/3305890.3306024}, our approach allows us to determine whether a frame's contribution is textual or acoustic. We hope this distinction paves the way for more granular and interpretable explanations within the speech processing domain.

\section{Learning to Disentangle} 
\label{sec:approach}

This section explains how we build upon the Information Bottleneck principle to disentangle textual and task-relevant acoustic features in speech representations.

\subsection{Variational Information Bottleneck}
The core idea of an Information Bottleneck (IB, \citealp{tishby2000information}) is to learn a stochastic encoding $Z$ that preserves maximal information about a target variable $Y$, while discarding irrelevant details of the input $H$. Accordingly, the loss function could be defined as: $\mathcal{L}_\text{IB} = - I(Z, Y) + \beta I(H, Z)$
where $I(., .)$ represents mutual information that measures the dependence between two variables. The coefficient $\beta \geq 0$ controls the trade-off between retaining information about either H or Y in Z.
This objective renders the IB framework especially attractive for disentanglement, and it has been effectively employed in diverse domains \citep{Qian2020UnsupervisedSD, Jeon2021IBGANDR, Dang2022DisentangledGW, wang2023visual}.
In practice, however, the exact computation of IB loss is intractable for continuous, high-dimensional variables because the mutual information terms require integration over unknown marginal distributions. To address this, the Variational Information Bottleneck (VIB, \citealp{alemi2016deep}) introduces the following efficient, variational bound\footnote{See \cite{alemi2016deep} for derivation.} to optimize:
\begin{equation}
\begin{aligned}
    \mathcal{L}_\text{VIB} &= \underbrace{\frac{1}{N} \sum^N_{n=1} \underset{z\sim p_\theta(z|h_n)}{\mathds{E}}[-\text{log}q_\phi(y_n|z)]}_{\text{Task loss}} \; + \\
    & \phantom{=\,} \beta \underbrace{\frac{1}{N} \sum^N_{n=1} \text{KL}[p_\theta(z|h_n), r(z)]}_{\text{Information loss}}
\label{eq:vib}
\end{aligned}
\end{equation}
\noindent where $N$ denotes the sample size; $p_\theta(z|h)$ serves as a stochastic encoder, mapping the input $h$ to the latent representation $z$; $q_{\phi}(y|z)$ acts as a parameterized decoder\footnote{Formally, this represents a variational approximation of the true $p(y_n |z)$, which is intractable as VIB assumes that the joint distribution factorizes as $p(y_n,z,h_n) = p(h_n) p(z | h_n) p (y_n | h_n)$~\citep{alemi2016deep}.}, a neural network that predicts the label $y$ from the latent representation $z$; and $r(z)$ approximates the marginal  $p(z)$. The first term in the loss function encourages the encoder to preserve information needed to predict the target label, while the second term, the KL divergence, pushes it to discard excessive information about $h$. The approximate marginal $r(z)$ is modeled as a spherical Gaussian \citep{alemi2016deep}, as it yields a closed-form solution for the KL divergence term.  The encoder $p(z|h)$ is parameterized using an MLP to predict the mean $\mu$ and the diagonal covariance matrix $\Sigma$: $p_{\theta}(z|h) = \mathcal{N}(z | \mu_{\theta}(h), \Sigma_{\theta}(h))$. 
Sampling from $p_{\theta}(z|h)$ is implemented using the reparameterization trick~\citep{kingma2013auto}, which enables joint optimization of $\theta$ and $\phi$ using stochastic gradient descent.

\begin{figure*}[t]
    \centering
    \includegraphics[width=0.9\textwidth]{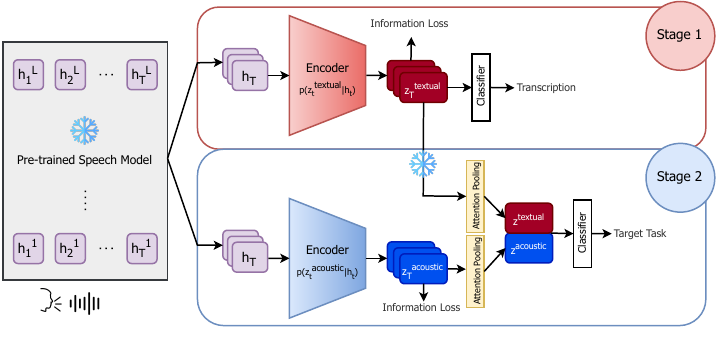}
    \caption{An overview of our cascaded setup: In stage 1 (text extraction), hidden states derived from a frozen pre-trained speech model are compressed using VIB to decode transcription. In stage 2 (acoustic extraction), the hidden states are compressed to predict the target task, conditioned on the frozen textual latent representations learned in stage 1. The training procedure results in lightweight textual and acoustic encoders that disentangle speech frame representations into \textcolor{red}{textual} and task-relevant \textcolor{blue}{acoustic} features.}
    \label{fig:diagram}
\end{figure*}

\subsection{Cascaded disentanglement}
\label{sec:proposed_method}
Consider a sequence of speech representations ($\bm h_1, ..., \bm h_T$), produced by a pre-trained speech model, where $\bm h_t$ represents the speech frame representation at timestamp $t$, and $T$ denotes the total number of frames in a given utterance. 
Given a target task, our goal is to decompose each frame representation into two distinct latent representations: \textcolor{red}{$\bm z^\text{textual}_t$} and \textcolor{blue}{$\bm z^\text{acoustic}_t$}. 
Our approach to disentangling these two latent representations is cascaded, and involves two stages (sketched in Figure~\ref{fig:diagram}) to minimize the Equation \ref{eq:vib}. Both stages operate on frozen speech representations from the same speech model.
We experiment with both weighted averages across all layers (Sec.~\ref{sec:results}- \ref{sec:evaluation}) and individual layer representations (Sec.~\ref{sec:layerwise_contribution}).

\subsubsection{Stage~1: text extraction}
In the first stage, we aim to distill the textual content from speech frame representations, while discarding other non-textual features.
The textual capability of a speech model is typically evaluated using automatic speech recognition (ASR), and measured by the word error rate (WER) metric, which assesses how accurately the model can transcribe spoken utterances based on their representations. 

We extract speech frame representations from a pre-trained model and compute a weighted average across its layers, with weights learned during training. These weights help maximize the model’s capability, as relevant information is not necessarily uniformly distributed across the model layers.
These frame representations are then given as input to a lightweight bottleneck encoder, which compresses the information into low-dimensional latent frame representations.
Next, we employ the Connectionist Temporal Classification (CTC, \citealp{10.1145/1143844.1143891}) loss as the task loss term in Equation \ref{eq:vib} to decode transcriptions from the latent frame representations, thus minimizing the following loss function:
\begin{equation}
    \mathcal{L}_{\text{Stage}_1} = \mathcal{L}_\text{CTC} + \beta \mathcal{L}_\text{Information}
\label{eq:stage_1}
\end{equation}
In this way, we force the bottleneck encoder $p(\bm z^\text{textual}|\bm h)$ to retain only the information necessary for transcription (as encouraged by the CTC loss), while discarding irrelevant features in the original representation ($\bm h$) (constrained by the information loss). 
We refer to the latent frame representation for frame $t$ learned at this stage as \textcolor{red}{$\bm z^\text{textual}_t$}.

\subsubsection{Stage~2: acoustic extraction}
Our goal in the second stage is to isolate acoustic features that complement the textual features learned in stage 1 and contribute to the target task. To achieve this, we replace the task loss in Equation \ref{eq:vib} with Cross-Entropy loss (CE) over the target class labels, thus, minimizing the following loss function:
\begin{equation}
    \mathcal{L}_{\text{Stage}_2} = \mathcal{L}_\text{CE} + \beta \mathcal{L}_\text{Information}
\label{eq:stage_2}
\end{equation}
We extract speech frame representations from the same speech model and again learn a weighted average over layers. These representations are then fed into a bottleneck encoder $p(\bm z^\text{acoustic}|\bm h)$ --- with the same architecture as the encoder in stage~1 --- to form the complementary latent representations (\textcolor{blue}{$\bm z^\text{acoustic}_t$}). 
We apply the information loss to the frame-wise latent representations at the output of the encoder. Subsequently, we pass these latent representations through an attention layer\footnote{We choose an attention layer over a simple average pooling to later recognize the key frame representations for interpretability.} to have latent representations at the utterance level, since labels for our target tasks are assigned to the whole utterance. 
Finally, the pooled latent representation \textcolor{blue}{$\bm z^\text{acoustic}$} is concatenated with the frozen textual latent representations \textcolor{red}{$\bm z^\text{textual}$} (previously trained in Stage~1) to decode the target task. This conditional setup encourages the trainable latent representations to avoid encoding textual information - since the decoder already has
access to it and the information minimization term
discourages redundancy. Instead, it is expected to
capture additional acoustic characteristics that are
beneficial for the target task.
During the training process in this stage, no gradient updates are directed back to the \textcolor{red}{$\bm z^\text{textual}_t$} already learned in stage~1. 

\subsubsection{Methodological considerations}
In an ablation experiment (Table~\ref{tab:sanity_no_ib}), we trained the encoders without the information loss. 
While removing the constraint yields a marginal improvement in raw performance, it leads to overlapping information between the two components, failing to achieve the intended disentanglement of textual and acoustic features (shown in Section~\ref{sec:evaluation}). This confirms the central role of the information loss in our setup.

It's noteworthy that the choice of the order of the stages is also not arbitrary: for tasks leveraging both textual and acoustic features, such as emotion recognition, training Stage 1 on the target task entangles acoustic features with textual ones, as the latter also contributes to the target task.

In a second ablation experiment (Table~\ref{tab:joint_vib}), we explored the feasibility of jointly training textual and acoustic encoders as an alternative to our two-stage approach. However, we found that joint training consistently led to considerable performance degradation across all tasks and models. Moreover, training in stages offers a key advantage: the textual latent representations learned in stage 1 are task-agnostic and can thus be trained once and reused across different downstream tasks. This makes the disentanglement pipeline highly efficient, as stage 2 can then be trained directly on the new target task within just a few minutes.

\section{Experimental setup}
\label{sec:experiments}
As our first case study, we choose emotion recognition as the target task, where both the content and acoustic features of spoken utterances are contributing.
For example, consider the utterances ``I'm so happy'' and ``I'm fine,'': where the former explicitly conveys happiness through its content, the latter can represent a wide range of emotions depending on prosody and tone. 
Our second case study closely follows the setup of the first, with speaker identification as the target task. For this task, we hypothesize that content offers only limited information, and the acoustic information will play an almost exclusive role in revealing the identity. 

\subsection{Training details}
For training the VIB, we follow \cite{alemi2016deep} and model $r(\bm z)$ and $p(\bm z|\bm h)$ as multivariate Gaussian distributions: $r(\bm z) = \mathcal{N}(\bm z|\bm \mu\!=\!\bm 0, \bm \Sigma\!=\!\bm 1)$ and $p(\bm z|\bm h)=\mathcal{N}(\bm z|\bm \mu(\bm h), \bm \Sigma(\bm h))$. 
The bottleneck encoders to estimate $\bm \mu$ and $\bm \Sigma$ consist of two shared linear layers with the same dimensionality as the original representations ($\bm h_t$), followed by independent $d$-dimensional layers for each, with GELU \citep{Hendrycks2016GaussianEL} activation functions in between. 
We experiment with different bottleneck dimensions $d=\{16, 32, 64, 128, 256\}$ as the output size of the bottleneck encoders.
To estimate the gradients, we employ the reparameterization trick \citep{kingma2013auto}: $\bm z_t=\bm \mu(\bm h_t) + \bm \Sigma(\bm h_t)\odot \bm \epsilon$, where $\bm \epsilon$ is sampled from the normal distribution $\mathcal{N}(\bm 0, \bm 1)$.
During inference, we use $\bm z_t=\bm \mu(\bm h_t)$.
During training, we gradually increase the $\beta$ coefficient (in Eq.~\ref{eq:vib}) linearly from $0.1$ to $1$. As shown in Appendix~\ref{app:beta_ablation}, gradually increasing $\beta$ is necessary and outperforms using a constant value, as it allows the network to first distill meaningful predictive features before applying drastic compression. The incremental schedule also eliminates the need for extensive hyperparameter search for $\beta$.
For decoding, we use a randomly initialized linear projection into $C$ classes, where $C\!=\!32$ for transcription (corresponding to the number of target characters), $C\!=\!4$ for emotion recognition, and $C\!=\!24$ for speaker identification. 

All models were trained for $50$ epochs using the AdamW optimizer with gradient norm clipping. The learning rate was set to $1\mathrm{e}{-3}$ and $1\mathrm{e}{-4}$ for Base and Large models, respectively, with a warmup ratio of $0.1$ and a cosine decay scheduler. For training transcription prediction using CTC loss, we use a batch size of $1$; the effective batch size here is the number of frame representations since the CTC loss is computed across all frames. For other training objectives, we use a batch size of $8$.

\begin{table*}[t]
    \centering
    \scriptsize  
    \begin{tabular}{l l | c g | c g | c g}
    \toprule[1pt]
    &  & \multicolumn{2}{c}{\textcolor{red}{Stage 1}} & \multicolumn{2}{c}{\textcolor{blue}{Stage 2}} & \multicolumn{2}{c}{\textcolor{blue}{Stage 2}} \\
     &  & \multicolumn{2}{c}{\textbf{Transcription} (WER $\downarrow$)} & \multicolumn{2}{c}{\textbf{Emotion} (Acc. $\uparrow$)} & \multicolumn{2}{c}{\textbf{Speaker Id.} (Acc. $\uparrow$)} \\ 
    \cmidrule(lr){3-4}
    \cmidrule(lr){5-6}
    \cmidrule(lr){7-8}
    \textbf{Model} & \textbf{Size} & \text{Original} & \text{VIB} & \text{Original} & \text{VIB} & \text{Original} & \text{VIB} \\ 
    \midrule[1pt]
    \multirow{2}{*}{HuBERT} & Base & $\text{0.45}_{\pm \text{0.000}}$ & $\text{0.41}_{\pm \text{0.006}}$ & $\text{0.62}_{\pm \text{0.007}}$ & $\text{0.63}_{\pm \text{0.006}}$ & $\text{0.98}_{\pm \text{0.001}}$ & $\text{0.99}_{\pm \text{0.001}}$ \\
    & Large & $\text{0.35}_{\pm \text{0.007}}$ & $\text{0.37}_{\pm \text{0.041}}$ & $\text{0.57}_{\pm \text{0.002}}$ & $\text{0.66}_{\pm \text{0.035}}$ & $\text{0.93}_{\pm \text{0.003}}$ & $\text{0.98}_{\pm \text{0.013}}$ \\
    \midrule
    \multirow{2}{*}{Wav2Vec2} & Base & $\text{0.50}_{\pm \text{0.000}}$ & $\text{0.45}_{\pm \text{0.002}}$ & $\text{0.62}_{\pm \text{0.001}}$ & $\text{0.59}_{\pm \text{0.033}}$ & $\text{0.99}_{\pm \text{0.000}}$ & $\text{0.98}_{\pm \text{0.017}}$ \\
    & Large & $\text{0.47}_{\pm \text{0.001}}$ & $\text{0.48}_{\pm \text{0.034}}$ & $\text{0.62}_{\pm \text{0.004}}$ & $\text{0.66}_{\pm \text{0.009}}$ & $\text{0.98}_{\pm \text{0.004}}$ & $\text{0.99}_{\pm \text{0.000}}$ \\
    \midrule
    HuBERT-FT & Large & $\text{0.03}_{\pm \text{0.000}}$ & $\text{0.08}_{\pm \text{0.077}}$ & $\text{0.55}_{\pm \text{0.013}}$ & $\text{0.65}_{\pm \text{0.040}}$ & $\text{0.90}_{\pm \text{0.003}}$ & $\text{0.98}_{\pm \text{0.003}}$ \\
    \midrule
    \multirow{2}{*}{Wav2Vec2-FT} & Base & $\text{0.06}_{\pm \text{0.000}}$ & $\text{0.12}_{\pm \text{0.006}}$ & $\text{0.63}_{\pm \text{0.006}}$ & $\text{0.60}_{\pm \text{0.056}}$ & $\text{0.99}_{\pm \text{0.000}}$ & $\text{0.99}_{\pm \text{0.002}}$ \\
    & Large & $\text{0.03}_{\pm \text{0.000}}$ & $\text{0.08}_{\pm \text{0.001}}$ & $\text{0.62}_{\pm \text{0.004}}$ & $\text{0.63}_{\pm \text{0.007}}$ & $\text{0.98}_{\pm \text{0.001}}$ & $\text{0.99}_{\pm \text{0.000}}$ \\
    \midrule
    \multirow{1}{*}{Whisper-enc} & Small & $\text{0.49}_{\pm \text{0.001}}$ & $\text{0.47}_{\pm \text{0.028}}$ & $\text{0.63}_{\pm \text{0.001}}$ & $\text{0.72}_{\pm \text{0.022}}$ & $\text{0.96}_{\pm \text{0.002}}$ & $\text{0.99}_{\pm \text{0.003}}$ \\
    \bottomrule
    \end{tabular}
    \caption{Performance of decoders trained on the \emph{original} hidden states \& the VIB latent vectors (\mbox{$d\!=\!128$}). Lower WER and higher Accuracy are better. Random baselines: $\text{WER}\!=\!1$ for transcription, $\text{Accuracy}\!=\!0.25$ for emotion, and $\text{Accuracy}\!=\!0.04$ for speaker identification. Scores are averaged (with $\pm$ standard deviation) over three independent runs with different random seeds.
    }
    \label{tab:vib_s1_s2_128_large}
\end{table*} 

\subsection{Target models}
To obtain speech representations, we use a range of pre-trained speech encoders: Wav2Vec2 \citep{baevski2020wav2vec} and HuBERT \citep{Hsu2021HuBERTSS} in their Base (12-Transformer layers, 768-hidden size) and Large (24, 1024) configurations as self-supervised encoders, and the encoder of Whisper-small (12, 768; \citealp{Radford2022RobustSR}) as a model trained in a supervised manner. All
models are obtained from the HuggingFace \citep{wolf-etal-2020-transformers} library. For self-supervised models we consider both pre-trained (on raw speech) and fine-tuned\footnote{The HuBERT model lacks a released fine-tuned checkpoint in the Base size.} (on transcribed speech) variants.
Self-supervised models learn speech representations through masked prediction. Wav2Vec2 employs a contrastive loss to identify the masked speech frame among distractors, while HuBERT uses k-means clustering to create prediction targets. Fine-tuned versions of these models are further trained on additional labeled speech data by optimizing a linear classifier using CTC loss to decode transcription.

\subsection{Data}
\paragraph{Transcription.} For training in stage 1, we use subsets of two widely used read speech corpora for ASR training: \mbox{LibriSpeech} \citep{7178964} and Mozilla's Common~Voice~17.0 \citep{Ardila2019CommonVA}. The former is derived from audiobooks, while the latter is recorded by contributors reading sentences displayed on a screen.
We randomly select 4,000 examples from each corpus, ensuring an equal representation of gender and speaker ID, with each sample having a maximum duration of 14 seconds. This results in 17.4 hours of transcribed speech training data. Similar to \citep{baevski2020wav2vec}, in our experiments, we found that a few hours of transcribed data is sufficient to train a linear classifier for decoding transcription from pre-trained speech representations.
For evaluation, we report model performance on the entire test-clean set of the LibriSpeech dataset.
Following \citep{baevski2020wav2vec}, we remove non-spoken special characters (e.g., commas and periods) from transcriptions, as these are not included in the target vocabulary. 

\paragraph{Target tasks.} For emotion data in stage 2, we use IEMOCAP \citep{busso2008iemocap} database as its emotion labels are influenced by both lexical content and acoustic properties in utterances.
Following prior research \citep{Li2021FusingAO, Li2022ExplorationOA}, we exclude utterances without transcripts and combine the \emph{Happy} and \emph{Excited} labels to form a 4-way classification task. We then undersample the dataset to balance emotion classes, resulting in 4,064 utterances ($\sim\!5$ hours of audio, with an average duration of 4.4 seconds per segment). Each utterance is assigned one emotion from the label set: \{\emph{Angry}, \emph{Happy}, \emph{Neutral}, \emph{Sad}\}.
For Speaker Identity as our second target task, we use a subset of Mozilla’s Common Voice 17.0 dataset\footnote{We also experiment with VoxCeleb \citep{Nagrani2017VoxCelebAL} dataset (see Appendix~\ref{app:more_dataset}).} \citep{Ardila2019CommonVA}, consisting of 4,000 training and 1,000 test samples, stratified by two genders (male and female) and 24 speaker identities.
All audio files in this study are resampled to 16 kHz to match the sampling rate used for the pre-training data of the target models.

\section{Task performance}
\label{sec:results}
In this section, we test the effectiveness of the training of the disentangled representations against the original entangled representations on the downstream tasks.  We will verify the disentanglement in the next section.
For comparison, we also train identically structured decoders that rely on the original hidden states ($h_t$); we label this condition as `Original'. 
The performance of these classifiers (which can be viewed as a repurposed form of {\it probing}; \citealp{Alain2016UnderstandingIL, Tenney_2019}) serves as a strong baseline, representing the performance achievable relying on the hidden states, without compressing them into latent representations.  

Table \ref{tab:vib_s1_s2_128_large} reports both VIB and Original performances, averaged over three runs with different random seeds, for transcription, emotion recognition, and speaker identification tasks. 
For the transcription task at stage 1, VIB generally stays close to the original performance\footnote{Note: In the VIB setup, unlike `Original', the transcription classifier is trained on stochastic latent representations ($\bm z_t=\bm \mu(\bm h_t) + \bm \Sigma(\bm h_t)\odot \bm \epsilon$).
This, coupled with joint optimization (as the textual vector $\bm z_t$ evolves during training), makes optimization more challenging and results in suboptimal performance when the classifier is later evaluated on the deterministic latent representation (i.e., the mean: $\bm z_t = \mu(\bm h_t)$).
In contrast, the classifier in Original setup is trained directly on fixed, deterministic representations.
This mismatch explains the slightly higher WER observed for VIB in the fine-tuned models. Importantly, when we train a new classifier on the final latent representations ($\bm z_t = \mu(\bm h_t)$), both setups yield nearly identical results, as confirmed by the results in Section~\ref{sec:evaluation}.} (WER), implying the success of VIB training in compressing essential information for transcription ($\text{WER}\!=\!1$ presents the performance of the random baseline).\footnote{We also evaluated the transcription performance using character error rate (CER), yielding a similar overall pattern.}
As expected, representations derived from fine-tuned models show better transcription performance compared to those from pre-trained models (for both VIB and Original) as they are specifically tuned for transcription.
The same success is evident in decoding the target tasks in stage 2, where we report the accuracy of both VIB and Original models for emotion recognition and speaker identification tasks (the random baseline accuracy is 0.25 and 0.04, respectively). 
Overall, the classifiers benefit more from the specialized, compressed representations than from the original hidden states, as VIB encourages retaining only task-relevant information, resulting in more robust representations.

We also found VIB performance consistent across various bottleneck dimensions (details omitted).
As shown in Appendix \ref{app:layer_weight_averaging}, both approaches indeed use the weights for layer averaging, as the target information is not uniformly distributed across the model layers.

\providecommand{\TextinTable}{\textcolor{red}{$\bm z^\text{textual}$}}
\providecommand{\AcousticEmotioninTable}{\textcolor{blue}{$\bm z^\text{acoustic}_\text{emotion}$}}
\providecommand{\AcousticSpeakerinTable}{\textcolor{blue}{$\bm z^\text{acoustic}_\text{speaker}$}}

\begin{table*}[th!]
\centering
\scriptsize
\begin{tabular}{l c l c c c c c}
\toprule
\textbf{Model} & \textbf{Size} & \textbf{Component} & \textbf{\textcolor{red}{ASR}} & \textbf{\textcolor{blue}{Mean Intensity}} & \textbf{\textcolor{blue}{Mean Pitch}} & \textbf{\textcolor{blue}{Gender}} & \textbf{\textcolor{blue}{Speaker Identity}} \\ 
& & & (WER $\downarrow$) & (Acc. $\uparrow$) & (Acc. $\uparrow$) & (Acc. $\uparrow$) & (Acc. $\uparrow$) \\
\midrule
Random baseline & \makecell[c]{--} & \makecell[c]{--} & \bf{1} & \bf{0.25} & \bf{0.25} & \bf{0.50} & \bf{0.04} \\ 
\cmidrule{1-8}
\multirow{8}{*}{HuBERT} & \multirow{4}{*}{Base} & Hidden State & 0.57 & 0.53 & 0.79 & 1 & 0.98 \\
\cmidrule{3-8}
 & & \TextinTable & 0.44 & \bf{0.28} & \bf{0.27} & \bf{0.51} & \bf{0.07} \\
 & & \AcousticEmotioninTable & \bf{1} & 0.38 & 0.42 & 0.75 & 0.41 \\
 & & \AcousticSpeakerinTable & \bf{1} & 0.56 & 0.79 & 1 & 0.99 \\ 
 \cmidrule{3-8}
 & \multirow{4}{*}{Large} & Hidden State & 0.46 & 0.47 & 0.72 & 0.99 & 0.92 \\
 \cmidrule{3-8}
 & & \TextinTable & 0.32 & \bf{0.26} & \bf{0.29} & \bf{0.55} & \bf{0.06} \\
 & & \AcousticEmotioninTable & \bf{1} & 0.38 & 0.51 & 0.84 & 0.52 \\
 & & \AcousticSpeakerinTable & \bf{1} & 0.55 & 0.79 & 0.99 & 0.99 \\
\midrule
\multirow{8}{*}{Wav2Vec2} & \multirow{4}{*}{Base} & Hidden State & 0.61 & 0.63 & 0.79 & 1 & 1 \\
\cmidrule{3-8}
 & & \TextinTable & 0.48 & \bf{0.25} & \bf{0.27} & \bf{0.55} & \bf{0.07} \\
 & & \AcousticEmotioninTable & \bf{1} & 0.39 & 0.44 & 0.77 & 0.40 \\
 & & \AcousticSpeakerinTable & \bf{1} & 0.55 & 0.78 & 0.99 & 0.99 \\ 
 \cmidrule{3-8}
 & \multirow{4}{*}{Large} & Hidden State & 0.56 & 0.56 & 0.80 & 1 & 0.98 \\
 \cmidrule{3-8}
 & & \TextinTable & 0.37 & \bf{0.26} & \bf{0.27} & \bf{0.52} & \bf{0.07} \\
 & & \AcousticEmotioninTable & \bf{1} & 0.34 & 0.38 & 0.69 & 0.26 \\
 & & \AcousticSpeakerinTable & \bf{1} & 0.55 & 0.79 & 1 & 1 \\
\midrule
\multirow{4}{*}{HuBERT-FT} & \multirow{4}{*}{Large} & Hidden State & 0.14 & 0.43 & 0.67 & 0.99 & 0.90 \\
\cmidrule{3-8}
 & & \TextinTable & 0.17 & \bf{0.24} & \bf{0.27} & \bf{0.51} & \bf{0.05} \\
 & & \AcousticEmotioninTable & \bf{1} & 0.36 & 0.42 & 0.76 & 0.32 \\
 & & \AcousticSpeakerinTable & \bf{1} & 0.53 & 0.78 & 1 & 0.98 \\ 
\midrule
\multirow{8}{*}{Wav2Vec2-FT} & \multirow{4}{*}{Base} & Hidden State & 0.17 & 0.60 & 0.80 & 1 & 1 \\
\cmidrule{3-8}
 & & \TextinTable & 0.20 & \bf{0.25} & \bf{0.28} & \bf{0.56} & \bf{0.06} \\
 & & \AcousticEmotioninTable & \bf{1} & 0.30 & 0.36 & 0.65 & 0.17 \\
 & & \AcousticSpeakerinTable & \bf{1} & 0.54 & 0.79 & 0.99 & 0.98 \\ 
 \cmidrule{3-8}
 & \multirow{4}{*}{Large} & Hidden State & 0.14 & 0.56 & 0.80 & 1 & 0.98 \\
 \cmidrule{3-8}
 & & \TextinTable & 0.15 & \bf{0.25} & \bf{0.27} & \bf{0.51} & \bf{0.06} \\
 & & \AcousticEmotioninTable & \bf{1} & 0.33 & 0.37 & 0.66 & 0.27 \\
 & & \AcousticSpeakerinTable & \bf{0.99} & 0.55 & 0.79 & 1 & 1 \\
\midrule
\multirow{4}{*}{Whisper-enc} & \multirow{4}{*}{Small} & Hidden State & 0.60 & 0.58 & 0.73 & 1.0 & 0.92 \\
\cmidrule{3-8}
 & & \TextinTable & 0.42 & \bf{0.27} & \bf{0.27} & \bf{0.53} & \bf{0.06} \\
 & & \AcousticEmotioninTable & \bf{1} & 0.39 & 0.42 & 0.78 & 0.30 \\
 & & \AcousticSpeakerinTable & \bf{1} & 0.55 & 0.80 & 1 & 1 \\ 
\midrule
\midrule
\multirow{4}{*}{FACodec} & \multirow{4}{*}{--} & \textcolor{red}{Content} & \bf{1} & \bf{0.25} & \bf{0.26} & \bf{0.51} & \bf{0.02} \\
 & & \textcolor{blue}{Prosody} & \bf{1} & \bf{0.24} & 0.31 & 0.70 & \bf{0.02} \\
 & & \textcolor{blue}{Timbre} & \bf{1} & 0.30 & \bf{0.27} & \bf{0.52} & \bf{0.04} \\
 & & \textcolor{blue}{Residual} & \bf{1} & \bf{0.26} & \bf{0.26} & \bf{0.50} & \bf{0.02} \\ 
\midrule
\multirow{3}{*}{SpeechSplit} & \multirow{3}{*}{--} & \textcolor{red}{Content} & \bf{1} & \bf{0.26} & 0.30 & \bf{0.54} & \bf{0.03} \\
 & & \textcolor{blue}{Pitch} & \bf{1} & \bf{0.20} & \bf{0.21} & \bf{0.48} & \bf{0.05} \\
 & & \textcolor{blue}{Rhythm} & \bf{1} & 0.39 & \bf{0.27} & 0.56 & \bf{0.03} \\
\bottomrule
\end{tabular}
\caption{Probing performances of latent representations ($d\!=\!128$) learned at stages 1 and 2, along with original hidden states derived from different pre-trained models, and components of existing disentanglement frameworks, for transcription (ASR) and a set of audio features (Mean Intensity, Mean Pitch, Gender, and Speaker Identity). For each model and task, components with scores close to the random baseline (i.e., least decodable) are highlighted in \textbf{bold}.}
\label{tab:sanity_check}
\end{table*}

\section{Evaluation of disentanglement}
\label{sec:evaluation}
Our training procedure (Sec.~\ref{sec:proposed_method}) yields two lightweight encoders that disentangle each speech frame $h_t$ into latent representations: \textcolor{red}{$\bm z^\text{textual}_t$} and \textcolor{blue}{$\bm z^\text{acoustic}_t$}. In this section, we investigate these latent representations to validate if they are truly disentangled by probing them for various types of textual and acoustic information.
Given an audio input, the goal is that \textcolor{red}{$\bm z^\text{textual}_t$}, which is specialized for text, should not encode any aspects of the acoustic characteristics of the audio. In contrast, \textcolor{blue}{$\bm z^\text{acoustic}_t$}, which is specialized for audio features, should not contain any information about the audio transcription. 

\subsection{Quantitative evaluation}
To assess textual capability, we train a linear probing classifier on frozen latent representations to predict their transcriptions.
To estimate acoustic capability, we train a set of probing classifiers followed by an attention layer on latent representations to predict various acoustic features at the utterance level, including Mean Intensity, Mean Pitch, Gender, and Speaker Identity.
For comparison. we also probe the original hidden states for the same objectives with the identically-structured classifiers.
We use a subset of Mozilla's Common Voice dataset \citep{Ardila2019CommonVA}, consisting of 4,000 training and 1,000 test samples, stratified by two genders (male and female)\footnote{The exclusion of other groups is due to the binary labeling in the dataset, rather than a choice by the authors.} and 24 speakers. We extract labels for intensity and pitch directly from the raw audio waveforms using the Parselmouth toolkit \citep{JADOUL20181}. We then discretize them into four equally sized buckets based on quantiles to form a four-way classification.

Figure~\ref{fig:eval_intro} compares transcription and Mean Intensity probing performances across components derived from Wav2Vec2-FT. Table~\ref{tab:sanity_check} reports complete results for all probing tasks, models, and sizes. For each model and task, components with scores close to the random baseline (i.e., least decodable) are highlighted in bold.
As we can see, acoustic latent representations (for both target tasks; \AcousticEmotioninTable and \AcousticSpeakerinTable exhibit no awareness of the textual content of the audio as their performance for ASR matches the random baseline ($\text{WER}\!=\!1$).
Conversely, textual latent representations are as effective as the original hidden states and -- for non-finetuned models -- even outperform them at decoding transcription.

Looking into predicting acoustic features (Mean Intensity, Mean Pitch, Gender, and Speaker Identity), textual latent representations (\TextinTable) consistently show random performance, suggesting no acoustic features are encoded within them. Acoustic latent representations, however, show substantial probing performance despite not having any explicit acoustic objective in their training at stage 2. Interestingly, acoustic latent representations for the task of speaker identification (\AcousticSpeakerinTable) are better at encoding acoustic features than those of emotion recognition (\AcousticEmotioninTable), which highlights the importance of the third axis in our desiderata: different downstream tasks may require different degrees of acoustic features.

Additionally, in contrast to \AcousticSpeakerinTable, acoustic latent representations for emotion (\AcousticEmotioninTable) do not match the performance of the original hidden states for acoustic features. For example, for the Wav2Vec2 Large, the probing performance for Speaker ID based on hidden states is 0.98, while it is 0.26 for \AcousticEmotioninTable (the random baseline is $1/24 \approx 0.04$). This disparity suggests that not all those acoustic features encoded in the hidden representations are crucial for recognizing emotion; thus, not all of those features were retained in stage 2. These findings could be particularly important in real-world scenarios where, for privacy protection, encoding acoustic features without precisely identifying the speaker is essential.  
The pattern of the probing results is similar across model sizes.

\subsubsection{Comparison with existing frameworks}
\label{related_work_comparison} 
FACodec \citep{Ju2024NaturalSpeech3Z} and SpeechSplit \citep{Qian2020UnsupervisedSD} are two recent speech disentanglement frameworks with complex training architectures that decompose speech signals into predefined components, regardless of the target downstream task. In contrast to our approach, they achieve this by reconstructing the input audio during their training procedures. This enables them to evaluate their method on style transfer audio conversion, but they do not assess whether there is still overlap in the information contained in their (disentangled) components. Our probing evaluation, in contrast, is applicable to all models, and allows us to compare the quality of the disentangled internal representations obtained with different approaches.
We therefore extract the embeddings of the disentangled components from the FACodec and SpeechSplit frameworks and similarly trained a linear classifier to probe them for various types of textual and acoustic information,
and include the results in Table~\ref{tab:sanity_check}.

We observe that the acoustic components derived from their frameworks perform poorly in decoding acoustic features when compared to our acoustic latent representations. 
Also, it is not possible to linearly decode the transcription (content) of the audio from their content vectors (in contrast to our textual latent representations).
Furthermore, the content vector derived from the SpeechSplit shows better performance than its acoustic components at predicting Mean Pitch, indicating contamination of their content vector with acoustic features. 

These findings suggest that while existing work on speech disentanglement has mainly focused on reconstructing input audio with different styles or speakers, the actual disentanglement of their components has not been properly evaluated. For instance: FACodec \citep{Ju2024NaturalSpeech3Z} uses an external ASR model (HuBERT-Large fine-tuned for ASR) to transcribe the reconstructed speech as a proxy measure for evaluating its content vectors at transcription. This approach does not accurately assess the quality of the content vector as the HuBERT model itself introduces a confounding variable that influences the reported WER. In contrast, our ASR probing method directly evaluates the degree of regularity in the content components with respect to the transcription labels.

\subsection{Qualitative evaluation}
We visualize the latent textual and acoustic representations to gain insight into how they have been encoded in the representation space for emotion recognition. We use the RAVDESS dataset \citep{livingstone2018ryerson} which contains only two identical statements spoken by 24 actors 
with 4 different emotions. This makes it ideal for this analysis, as the linguistic content remains constant across utterances but with different emotions.
We select examples with matching emotion labels from the IEMOCAP dataset, resulting in 384 utterances. 
We obtain the speech representations from the large Wav2Vec2 model and generate their corresponding latent representations using the bottleneck encoders trained in stages 1 and 2, without any further training.
Then, we compute the average of frame representations over all frames in each utterance and apply t-SNE.

\begin{figure}[t]
    \centering
    \includegraphics[width=\columnwidth]{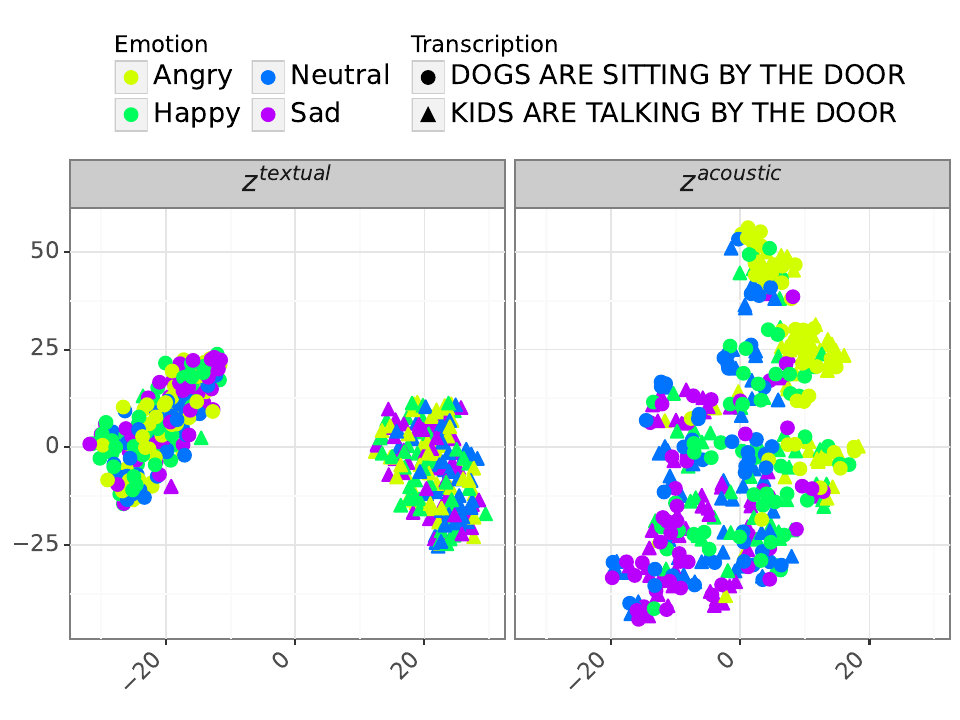}
    \caption{t-SNE of the textual (left) and acoustic (right) latent representations for Wav2Vec2-Large, marked and colored according to their transcription and emotion labels, respectively.}
    \label{fig:tsne_128_large}
\end{figure}

Figure \ref{fig:tsne_128_large} shows a 2D projection of these latent representations with data points marked and colored according to their transcription and emotion labels, respectively.
The textual representations 
(learned in stage 1) 
are clustered according to their transcriptions. In contrast, the acoustic representations 
(learned in stage 2) 
are not separated by transcription, suggesting that textual information is not retained there. Instead, these acoustic representations are roughly clustered by emotion, which aligns with our desired goal.
The pattern is similar for the Base model (not shown).

\begin{figure*}[th]
    \centering
    \begin{minipage}{0.33\textwidth}
        \centering
        \includegraphics[width=\linewidth]{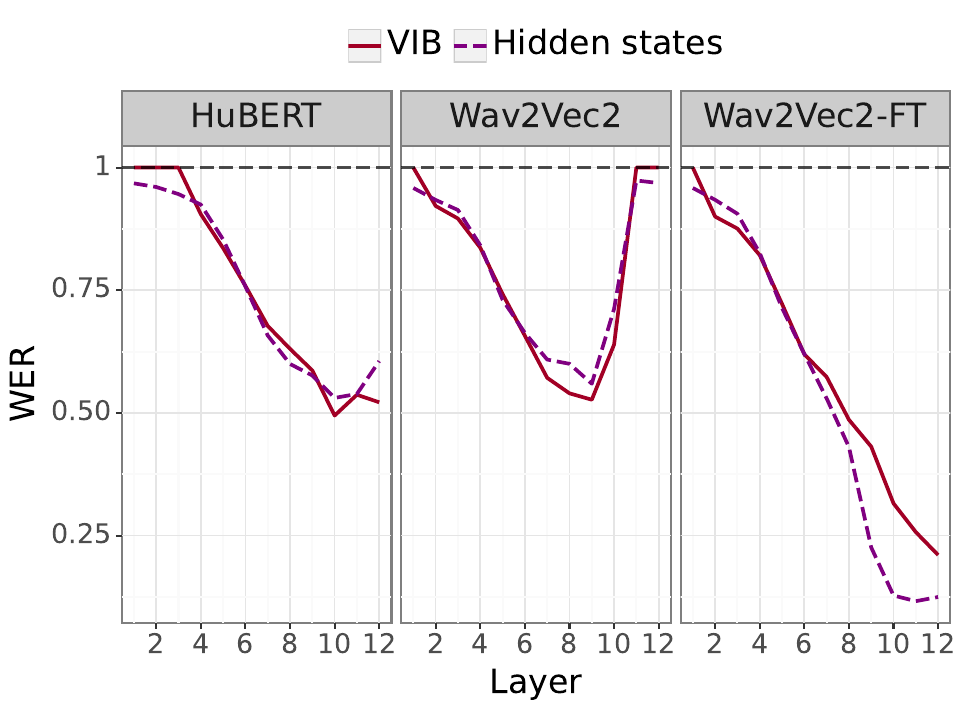}
    \end{minipage}%
    \begin{minipage}{0.33\textwidth}
        \centering
        \includegraphics[width=\linewidth]{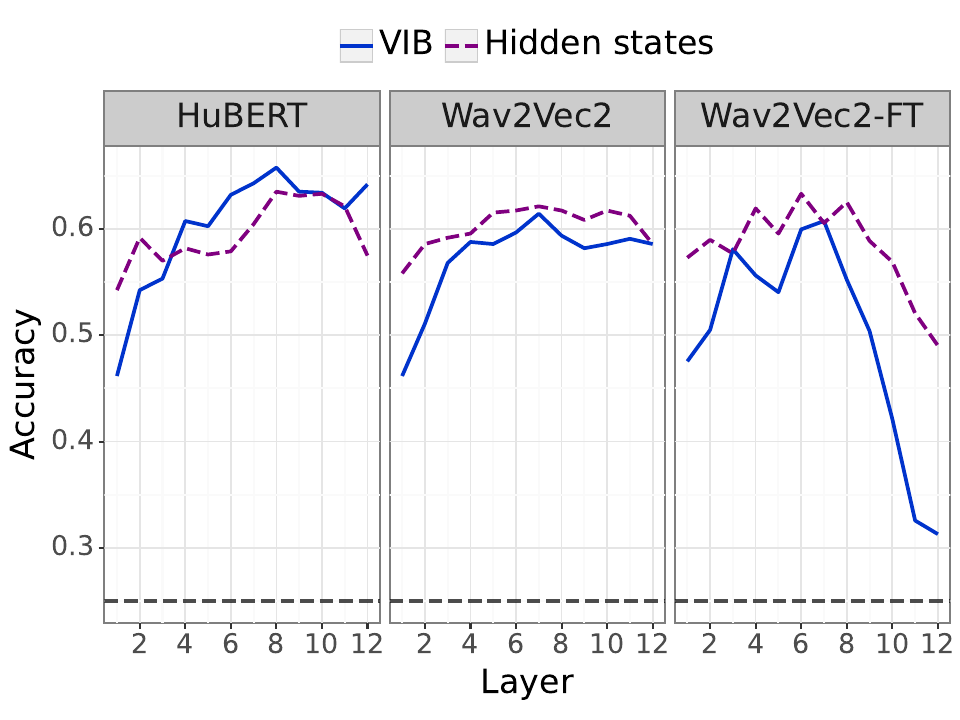}
    \end{minipage}%
    \begin{minipage}{0.33\textwidth}
        \centering
        \includegraphics[width=\linewidth]{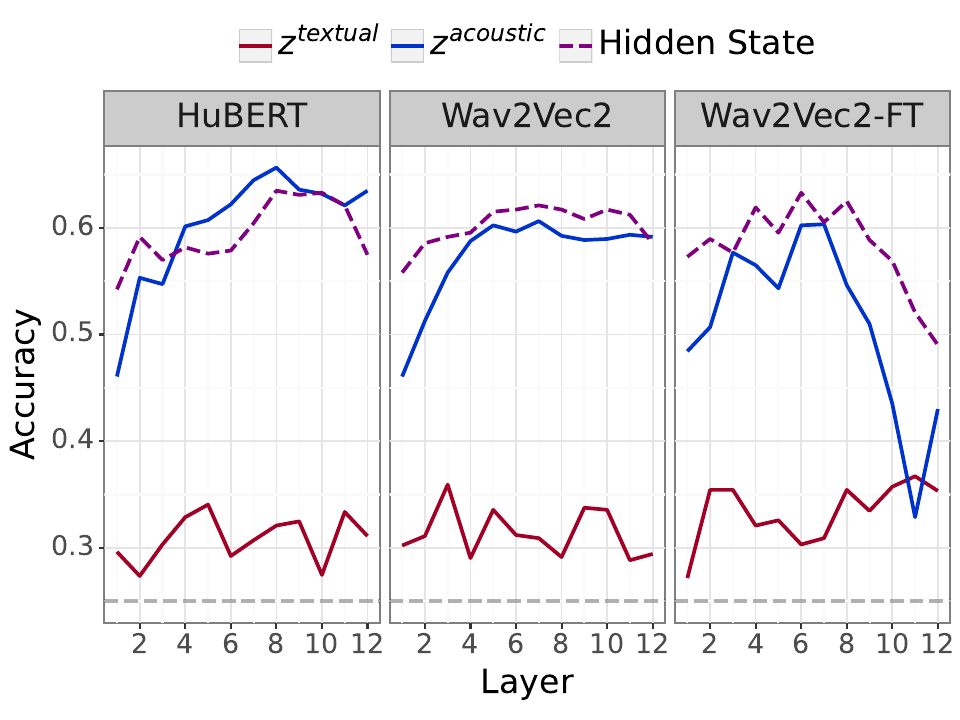}
    \end{minipage}
    \caption{Layerwise performance of VIB (\mbox{$d\!=\!128$}) for transcription at stage 1 ({\em left}) and emotion classification at stage 2 ({\em middle}), compared to layerwise performance using original hidden state activations. {\em Right}: layerwise textual and acoustic contribution to emotion classification, compared to layerwise performance using hidden states of Base models. Lower WER and higher accuracy are better. The black horizontal dashed line represents the random baseline.}
    \label{fig:layerwise_contribution}
\end{figure*}

\section{Using disentanglement for improved interpretability}
\label{sec:using}

\subsection{Disentangled probing}
\label{sec:layerwise_contribution}
With our approach to learning disentanglement validated, we can now quantify, separately, the extent to which each layer in a pre-trained speech model contributes textually and acoustically to the target task of emotion recognition.

We first train the latent representations using the same setup as we described in Section \ref{sec:proposed_method}, but instead of using a weighted layer average in the input, we use the frame representations from a specific layer of the model. The layerwise results for stages 1 and 2 are shown in Figure \ref{fig:layerwise_contribution}, with the black horizontal dashed line representing random performances.

For HuBERT, the transcription ability improves as we move through the layers. In contrast, Wav2Vec2 shows a U-shaped pattern, with the best performance in the middle layers and the final layers being unable to decode transcription. For Wav2Vec2-FT, however, transcription performance improves sharply in the last layers, which is expected given that the model is fine-tuned for ASR, confirming previous findings on substantial changes in the last layers of the model during fine-tuning \citep{Pasad2021LayerWiseAO}. In stage 2 (middle panel), models perform best at emotion recognition in the middle layers. Compared to pre-trained, the fine-tuned model shows a decline in emotion recognition performance in the final layers, where acoustic information is lost in favor of encoding text transcription.

We next train a probing classifier (followed by an attention layer) on top of frozen \textcolor{red}{$\bm z^\text{textual}_t$} and \textcolor{blue}{$\bm z^\text{acoustic}_t$} latent representations to decode emotion. Each latent representation is specialized to preserve either textual or acoustic features from the original hidden states of each model layer. Therefore, the probing performance reveals the extent to which these textual and acoustic features contribute to emotion recognition 
For comparison, we also train the same probing classifier on the original hidden states across each layer (similar to \citealp{Li2022ExplorationOA}; Figure \ref{fig:layerwise_contribution}, right).
Overall, both textual and acoustic components are beneficial, with their contributions varying across layers.
Comparing pre-trained and fine-tuned models, the latent acoustic representations learned from the final layers of Wav2Vec-FT contribute significantly less to emotion recognition. This is likely due to the model losing some acoustic information during fine-tuning in favor of transcription capabilities. However these layers benefit more from textual features in predicting emotion (which cannot be captured without disentanglement), as their representations offer more accurate transcriptions.
Compared to Wav2Vec2, HuBERT shows a greater contribution to emotion recognition; acoustically in the middle layers, and textually in the last layer.

\subsection{Disentangled feature attribution}
\label{sec:feature_attribution}

In this section, we use our disentangled vectors to localize salient input features for the target tasks. The attention layer in stage~2 (see Figure \ref{fig:diagram}) can be used to identify those frames in the original input audio whose latent representations contribute most to our target tasks.
The disentanglement mechanism allows us to clearly separate the contributions of acoustic features from those of textual features. This \emph{disentangled} attribution relates to \citep{wang2023visual}, which applies a multi-modal information bottleneck approach to identify relevant visual and textual features in vision-language models, and could be particularly useful in fields like psychiatry, where differences between textual and acoustic emotional expressions can aid in the diagnosis of disorders \citep{niu2023capturing}, or in detecting bias in speech agents' responses to user requests.

Finding salient input features for model decisions is often done by computing the gradient of the model's output with respect to the inputs \citep{10.5555/3305890.3306024, ancona2018towards, yuan2019interpreting, samek2019explainable, fucci-etal-2024-explainability}.
To compare our attention scores with gradient-based attribution scores, we train a classifier for the target task on the original hidden states of the Wav2Vec2 model and use Integrated Gradients (IG) to compute attribution to individual frames \citep{10.5555/3305890.3306024}. We normalize the IG scores to sum to $1$, ensuring a fair comparison with attention scores. 
This gives us three forms of attributions: acoustic attention, textual attention, and IG scores.

We then obtain frame-by-frame distributions of both acoustic and sentiment features to compare them directly with attribution results. For acoustic features, we focus on intensity and pitch, identifying and representing peaks and valleys in their temporal patterns. To analyze sentiment, we use the spaCy toolkit \citep{spacy2} to annotate word-level polarity (positive, negative, or neutral) within the utterances.  Using Montreal Forced Aligner \citep{mcauliffe17_interspeech}, we map these word-level labels to frames (see details in Appendix \ref{app:hyperparams}). As a result, each feature vector for a speech frame includes the following dimensions: (1) the presence of a peak or valley in intensity, (2) the same for pitch, and (3) the presence of a `sentiment-laden' word, all normalized to the range $[0, 1]$.

\begin{table}[t!]
\centering
\resizebox{1.0\columnwidth}{!}{%
\begin{tabular}{lccc}
\toprule
\textbf{Method} & \textbf{Intensity} & \textbf{Pitch} & \textbf{Sentiment} \\
\midrule
\textcolor{red}{Textual} Attention & 0.15 & 0.07 & \bf{0.15} \\
\textcolor{blue}{Acoustic} Attention & \bf{0.41} & \bf{0.15} & 0.05 \\
Integrated Gradient & 0.20 & 0.07 & 0.05 \\
\bottomrule
\end{tabular}
}
\caption{The alignment (Dot product) of attribution scores and different acoustic and textual cues.}
\label{tab:attribution_dot_product}
\end{table}

We then compute the dot product between the frame-wise attribution scores and the corresponding feature vectors. 
Table \ref{tab:attribution_dot_product}  presents these results, averaged across all examples in the IEMOCAP test set, and demonstrates that acoustic attention effectively captures peaks and valleys in acoustic features  (intensity and pitch), while textual attention focuses on word polarity. Both have higher agreement with features than Integrated Gradient scores.
Note that textual attention exhibits fairly high similarity with acoustic features as polar words are often pronounced with emotional emphasis.

\section{Conclusion and future work}
\label{sec:conclusions}

We present an approach to post-hoc disentanglement of representations from pre-trained speech models into textual and acoustic components, retaining only features relevant to the target tasks.    
Through a series of experiments on several pre-trained models with varying sizes and architectures, we show that our setup using the variational information bottleneck and cascaded learning can effectively isolate key features and improve interpretability while maintaining the performance of the original models, and can identify the most important speech frames from both textual and acoustic perspectives. 

The approach has potential for applications 
where disentanglement is key, such as speech AI systems, where privacy-preservation is key for responsible deployment. 
We want to guarantee that information about the speaker's identity, gender, or health characteristics is not used in downstream applications. 
Since these features stem from acoustic aspects of the audio, disentanglement allows us to suppress such information.
A unique aspect of our approach is its modality-independent architecture: in principle, any entangled multimodal representation could be disentangled as long as appropriate target tasks are implemented. This opens up applications beyond neural speech recognition, e.g.\ for disentangling linguistic and visual components in representations of grounded language models. 
We leave further exploration of these possibilities for future work.


\section*{Acknowledgments}
We would like to thank the TACL reviewers and action editors for their helpful suggestions.
Support from the Netherlands Organization for Scientific Research (NWO), through the National Research Agenda project ``InDeep: Interpreting Deep Learning Models for Text and Sound'' (NWA-ORC NWA.1292.19.399) and the VICI project ``Language-Driven Modularity in Neural Networks'' (VI.C.212.053) is gratefully acknowledged.
We also thank SURF (\url{www.surf.nl}) for the support in using the National Supercomputer Snellius. The work was partially done while Hosein was visiting the University of Edinburgh. 

\bibliography{tacl2021}
\bibliographystyle{acl_natbib}

\appendix 

\section{Appendix}

\subsection{Ablation experiments}
\label{app:ablation} 
\subsubsection{Ablation 1: Information bottleneck loss} 
To verify the importance of the Information Bottleneck, we trained the textual and acoustic encoders without the information loss (Equation \ref{eq:vib}). As shown in Table \ref{tab:sanity_no_ib}, excluding information loss during training results in a failure to disentangle textual and acoustic information as intended: for instance, the textual latent representations (learned in stage 1) can predict acoustic features even better than the acoustic representations themselves (learned in stage 2). This ablation confirms the central role of information loss in our approach.

\providecommand{\TextinTable}{\textcolor{red}{$\bm z^\text{textual}$}}
\providecommand{\AcousticinTable}{\textcolor{blue}{$\bm z^\text{acoustic}$}}
\providecommand{\TextWOIBinTable}{\textcolor{red}{$\bm z^\text{textual w/o}$}}
\providecommand{\AcousticWOIBinTable}{\textcolor{blue}{$\bm z^\text{acoustic w/o}$}}

\begin{table*}[th!]
\centering
\footnotesize
\begin{tabular}{l l c c c c c}
\toprule
\textbf{Model} & \textbf{Component} & \textbf{\textcolor{red}{ASR}} & \textbf{\textcolor{blue}{Mean Intensity}} & \textbf{\textcolor{blue}{Mean Pitch}} & \textbf{\textcolor{blue}{Gender}} & \textbf{\textcolor{blue}{Speaker Identity}} \\ 
& & (WER $\downarrow$) & (Acc. $\uparrow$) & (Acc. $\uparrow$) & (Acc. $\uparrow$) & (Acc. $\uparrow$) \\
\midrule
Random & \makecell[c]{--} & \bf{1} & \bf{0.25} & \bf{0.25} & \bf{0.50} & \bf{0.04} \\ 
\cmidrule{1-7}
\multirow{6}{*}{Wav2Vec2} & Hidden State & 0.61 & 0.63 & 0.79 & 1 & 1 \\
\cmidrule{2-7}
& \TextinTable & 0.48 & \bf{0.25} & \bf{0.27} & \bf{0.55} & \bf{0.07} \\
& \AcousticinTable & \bf{1} & 0.39 & 0.44 & 0.77 & 0.40 \\
\cmidrule{2-7}
& \TextWOIBinTable & 0.34 & \underline{0.39} & \underline{0.65} & \underline{0.91} & \underline{0.90} \\
& \AcousticWOIBinTable & \bf{1} & 0.37 & 0.44 & 0.73 & 0.37 \\
\bottomrule
\end{tabular}
\caption{Probing performances of latent representations learned at stages 1 and 2 \textbf{with and without information loss constraint} (marked by w/o), along with hidden states derived from Wav2Vec2-Base model. For each model and task, components with scores close to the random baseline (i.e., least decodable) are highlighted in \textbf{bold}. Conversely, scores that are expected to be random but exhibit non-random behavior (indicating overlapping information) are \underline{underlined}.}
\label{tab:sanity_no_ib}
\end{table*}

\begin{table*}[th!]
    \centering
    \resizebox{1.0\textwidth}{!}{%
    \begin{tabular}{l l | g c c | g c | g c}
    \toprule[1pt]
     &  & \multicolumn{3}{c}{\textbf{Transcription (WER $\downarrow$)}} & \multicolumn{2}{c}{\textbf{Emotion (Acc. $\uparrow$)}} & \multicolumn{2}{c}{\textbf{Speaker Id. (Acc. $\uparrow$)}} \\ 
    \cmidrule(lr){3-5}
    \cmidrule(lr){6-7}
    \cmidrule(lr){8-9}
    \textbf{Model} & \textbf{Size} & \shortstack{Cascaded VIB \\ (Stage 1)} & \shortstack{Joint VIB \\ (w/ Emotion)} & \shortstack{Joint VIB \\ (w/ Speaker Id.)} & \shortstack{Cascaded VIB \\ (Stage 2)} & \shortstack{Joint VIB \\ \phantom{(xxxxx x)}} & \shortstack{Cascaded VIB \\ (Stage 2)} & \shortstack{Joint VIB \\ \phantom{(xxxxx x)}} \\ 
    \midrule[1pt]
    \multirow{2}{*}{HuBERT} & Base & 0.41 & 0.64 & 0.92 & 0.63 & 0.62 & 0.99 & 0.98 \\
    & Large & 0.37 & 0.35 & 0.44 & 0.66 & 0.59 & 0.98 & 0.98 \\
    \midrule
    \multirow{2}{*}{Wav2Vec2} & Base & 0.45 & 1 & 1 & 0.59 & 0.56 & 0.98 & 0.95 \\
    & Large & 0.48 & 0.49 & 0.86 & 0.66 & 0.59 & 0.99 & 0.99 \\
    \midrule
    HuBERT-FT & Large & 0.08 & 0.16 & 0.33 & 0.65 & 0.64 & 0.98 & 0.88 \\
    \midrule
    \multirow{2}{*}{Wav2Vec2-FT} & Base & 0.12 & 0.17 & 0.74 & 0.60 & 0.54 & 0.98 & 0.97 \\
    & Large & 0.08 & 0.16 & 0.39 & 0.63 & 0.58 & 0.99 & 0.98 \\
    \end{tabular}
    }
    \caption{Comparison of \textbf{cascaded VIB} with \textbf{joint VIB}. 
    Random baselines: $\text{WER}\!=\!1$ (transcription), $\text{Accuracy}\!=\!0.25$ (emotion),  $\text{Accuracy}\!=\!0.04$ (speaker identification).}
    \label{tab:joint_vib}
\end{table*}

\subsubsection{Ablation 2: Cascaded vs. joint optimization setup} 
To assess the importance of cascading (sec. \ref{sec:proposed_method}), we run experiments where textual and acoustic latent variables are trained jointly. Specifically, at each training step, we feed the transcribed data into the textual encoder to calculate the CTC loss ($\mathcal{L}_\text{CTC}$), and feed the acoustic encoder with target task data to compute the cross-entropy loss ($\mathcal{L}_\text{CE}$). Both latent variables are also constrained by information bottleneck ($\mathcal{L}_\text{Info}$) losses. Then, we combine losses and perform backpropagation to jointly optimize both model parameters to minimize:
\begin{equation}
    \mathcal{L}_\text{CTC} - \beta \mathcal{L}_{\text{Info}_1} + \mathcal{L}_\text{CE} - \beta \mathcal{L}_{\text{Info}_2}
\end{equation}

Incorporating transcribed data alongside the target task is necessary during the joint training because: 1) we have limited transcribed data for target downstream tasks (e.g., ~5 hours for emotion recognition), which is not sufficient for learning transcription, and 2) in real-world applications, transcriptions of speech are not always available. In this way, both classifiers are trained on the same data as in our two-stage cascaded training, but now they are jointly optimized, making them comparable.

Table \ref{tab:joint_vib} 
shows that joint training consistently and significantly underperforms the cascaded approach, and in two cases (for pre-trained Wav2Vec2), it completely fails to decode transcriptions, which highlights the importance of our cascaded setup.
Moreover, cascaded training offers key advantages: the textual
latent representations in stage 1 can be trained once and reused
across all target tasks, while stage 2 can be trained efficiently
in just a few minutes.

\subsection{The impact of $\beta$}
\label{app:beta_ablation}

\begin{figure}[ht]
    \centering
    \begin{minipage}{\columnwidth}
        \centering
        \includegraphics[width=0.9\columnwidth]{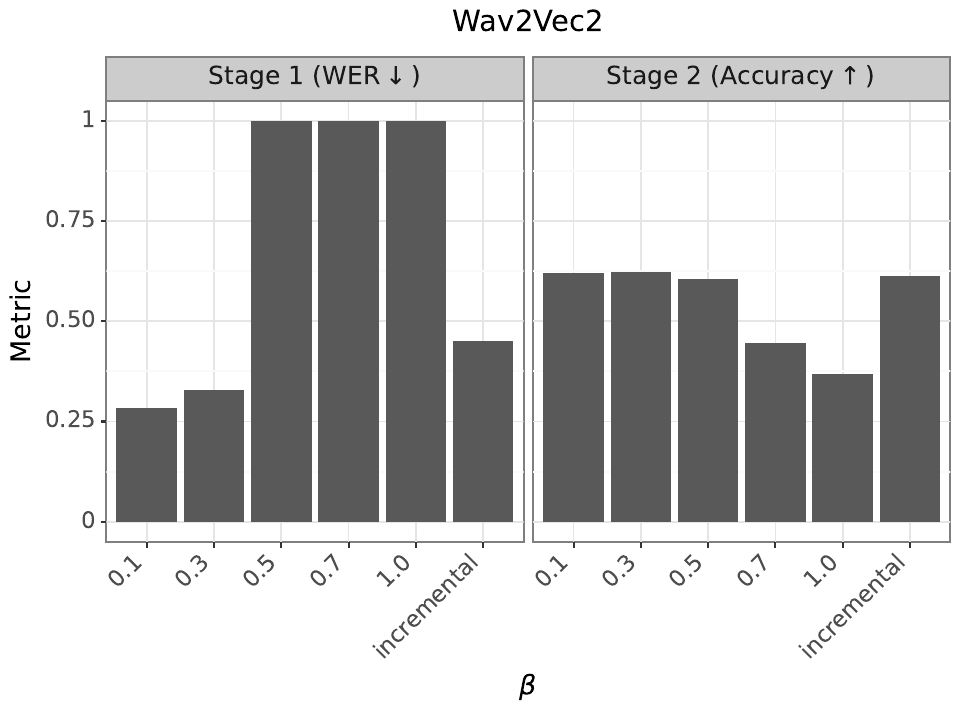}
    \end{minipage}
    \begin{minipage}{\columnwidth}
        \centering
        \includegraphics[width=0.9\columnwidth]{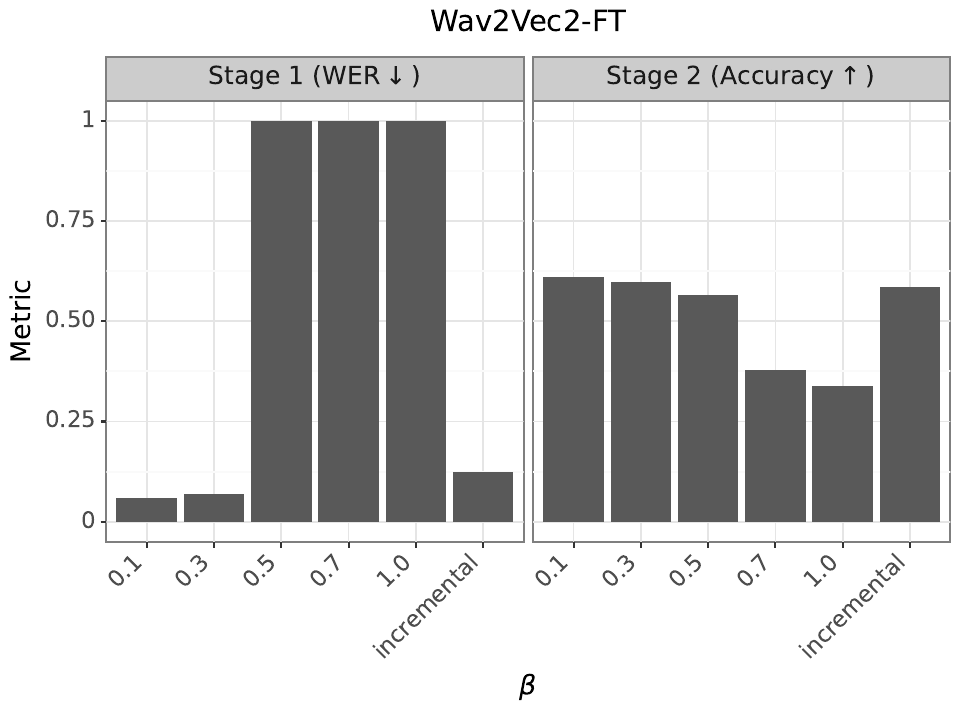}
    \end{minipage}
    \caption{The effect of varying constant $\beta$ values compared to incremental $\beta$ schedule on training VIB for transcription (stage 1) and emotion (stage 2).}
    \label{fig:beta_ablation}
\end{figure}

In the VIB loss (Eq. \ref{eq:vib}), the coefficient $\beta$ controls the trade-off between compression and prediction: higher values force the latent representation to forget more of the original hidden states by strongly penalizing mutual information, while lower values relax this constraint, allowing it to retain more information for predicting the target task.
During training, we gradually increase $\beta$ linearly from $0.1$ to $1$. In this section, we perform an ablation study to validate this choice by comparing it against training with fixed $\beta$ values.

Figure~\ref{fig:beta_ablation} presents task performance for VIB applied to the Wav2Vec2 and Wav2Vec2-FT models under different values of $\beta$. Performance is reported for transcription in Stage~1 and emotion recognition in Stage~2, with the same $\beta$ value used in both stages.
While low constant values of $\beta \in \{0.1, 0.3\}$ can achieve good task performance, they are insufficient for effective disentanglement: the mutual information regularization is too weak to suppress task-irrelevant information, resulting in overlap among latent components. In contrast, using a high fixed $\beta \in \{0.5, 0.7, 1\}$ over-regularizes the representations and destabilizes optimization, preventing convergence and yielding random WER in stage~1 and poor emotion recognition accuracy in stage~2, across all models.
The proposed incremental schedule strikes a balance: it allows the model first to capture predictive features and then progressively enforces compression. Moreover, it substantially reduces sensitivity to the precise choice of $\beta$, alleviating the need for exhaustive hyperparameter tuning.

\begin{figure}[t]
    \centering
    \begin{minipage}{\columnwidth}
        \centering
        \includegraphics[width=\columnwidth]{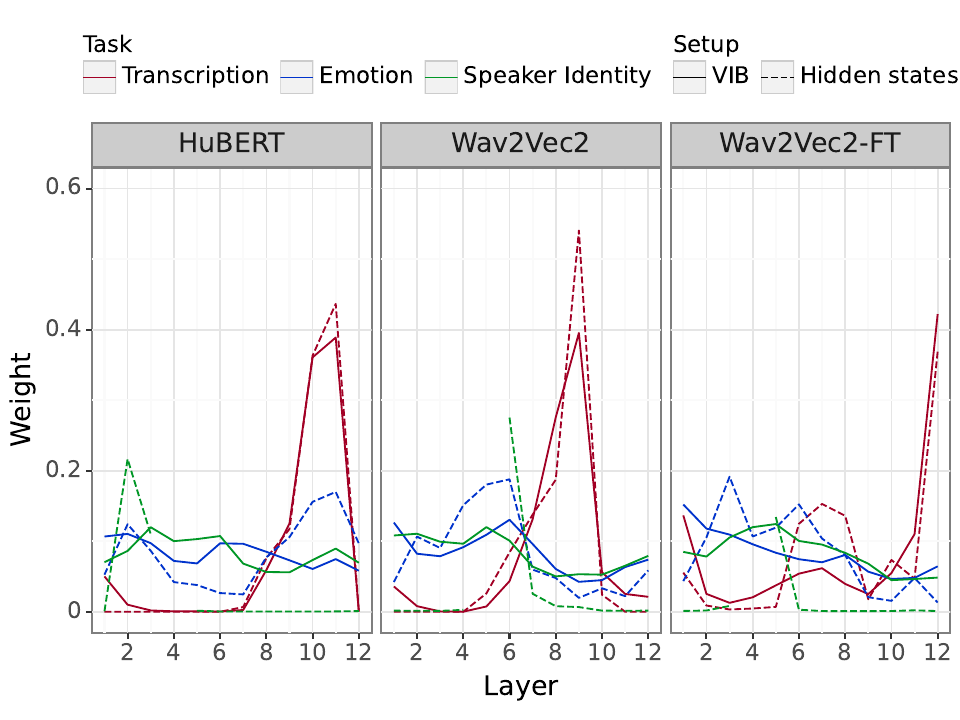}
    \end{minipage}
    \begin{minipage}{\columnwidth}
        \centering
        \includegraphics[width=\columnwidth]{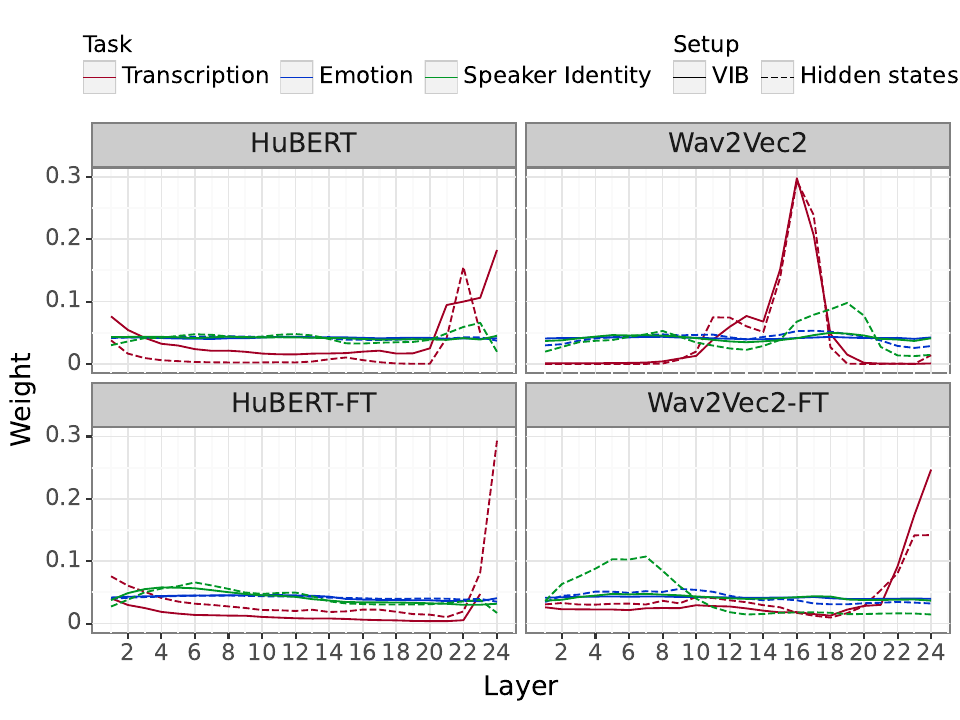}
    \end{minipage}
    \caption{Layer weights of \textbf{Base} (top) and \textbf{Large} (bottom) models, trained for transcription (stage 1), emotion and speaker identity (stage 2), compared with probes applied to the original hidden states.}
    \label{fig:weights_128}
\end{figure}

\subsection{Layerwise weights}
\label{app:layer_weight_averaging}

In the input of both VIB and original training setups in the experiments in Section~\ref{sec:results}, we train weights for layer averaging as information for transcription, emotion, and speaker identification is not uniformly distributed over model layers.
Figure \ref{fig:weights_128} shows these trained layer weights for both Base (top) and Large (bottom) models.
Overall, layer weights learned in the VIB setup closely follow the weights trained with original hidden states in the probing setup.

In stage 1, pre-trained models rely on upper-middle layers for transcription, while fine-tuned ASR models shift this to the final layer, consistent across Base and Large models and different bottleneck sizes. For stage 2 tasks (Emotion and Speaker Identification), Base models rely more on early layers, whereas Large models show uniform contribution, suggesting that the acoustic information needed for these tasks is distributed across layers.

\begin{table}[ht!]
    \centering
    \tiny
    \resizebox{0.9\columnwidth}{!}{%
    \begin{tabular}{l | c g}
    \toprule
    & \multicolumn{2}{c}{\textcolor{blue}{Stage 2}} \\
    & \multicolumn{2}{c}{\textbf{Speaker Identification} (Acc. $\uparrow$)} \\ 
    \cmidrule(lr){2-3}
    \textbf{Model} & \text{Original} & \text{VIB} \\ 
    \midrule
    HuBERT & 0.70 & 0.87 \\
    \midrule
    Wav2Vec2 & 0.57 & 0.58 \\
    \end{tabular}
    }
    \caption{Performance of linear decoders trained on the \emph{original} hidden representations and on the VIB latent vectors ($d = 128$) for Speaker Identification on the \textbf{VoxCeleb} dataset. The random baseline accuracy for the 40-way classification task is $0.025$.
    }
    \label{tab:stage_2_vox}
\end{table} 

\subsection{More datasets for speaker identification}
\label{app:more_dataset}
In this section, we perform disentanglement on the VoxCeleb \citep{Nagrani2017VoxCelebAL} dataset, as an alternative dataset, when the target task is speaker identification. VoxCeleb is an open-set corpus preliminary purposed for speaker verification. To adapt it for speaker identification, we construct a subset of the training split containing 40 distinct speakers.
From this subset, we create train and test partitions such that both splits contain the same speaker identities but non-overlapping audio utterances, resulting in 1,200 training samples and 400 test samples.
We reuse the pre-trained textual encoder and train only Stage 2 on the target VoxCeleb speaker identification task.
Table \ref{tab:stage_2_vox} reports classification accuracy for both the Original and VIB setups for pre-trained HuBERT and Wav2Vec2 models. For reference, the random-guess baseline is $1/40=0.025$.
Overall, the results demonstrate that the proposed cascaded approach generalizes effectively to different datasets.

\subsection{Temporal alignment}
\label{app:hyperparams}
To map the frame representations of the encoder model to their corresponding audio timestamp and transcriptions, we use the Montreal Forced Aligner \citep[MFA]{mcauliffe17_interspeech}. Using this tool, we extract the start time ($t_s$) and the end time ($t_e$) of each word in an utterance, and map them to boundary frames $f_s$ and $f_e$:
$f = \lceil \frac{t}{\mathcal{T}} \times T \rceil$
\noindent where $\mathcal{T}$ and $T$ denote the total time of a given audio and the total number of frames in the frame representation, respectively.

\end{document}